# Intelligent decision-making method of TBM operating parameters based on multiple constraints and objective optimization

**Author information**

Bin Liu[a,b,c*], Jiwen Wang[a,b], Ruirui Wang[d,a], Yaxu Wang[a,e], Guangzu Zhao[a,f]

[a] Geotechnical and Structural Engineering Research Center, Shandong University, Shandong, China

[b] School of Civil Engineering, Shandong University, Shandong, China

[c] Data Science Institute, Shandong University, Shandong, China

[d] School of Civil Engineering, Shandong Jianzhu University, China

[e] School of Qilu Transportation, Shandong University, Shandong, China

[f] Zhejiang Huadong Construction Engineering Co., Ltd, Zhejiang, China

**\*Corresponding author:**

Bin Liu, corresponding at Geotechnical and Structural Engineering Research Center, Shandong University, No. 17923 Jingshi Road, Post 250061, Jinan, China.

E-mail address: liubin0635@163.com

**Abstract**

The decision-making of TBM operating parameters has an important guiding significance for TBM safe and efficient construction, and it has been one of the research hotpots in the field of TBM tunneling. For this purpose, this paper introduces rock-breaking rules into machine learning method, and a rock-machine mapping dual-driven by physical-rule and data-mining is established with high accuracy. This dual-driven mappings are subsequently used as objective function and constraints to build a decision-making method for TBM operating parameters. By searching the revolution per minute and penetration corresponding to the extremum of the objective function subject to the constraints, the optimal operating parameters can be obtained. This method is verified in the field of the Second Water Source Channel of Hangzhou, China, resulting in the average penetration rate increased by 11.3%, and the total cost decreased by 10.0%, which proves the practicability and effectiveness of the developed decision-making model.

**Keywords:** TBM operating parameters, rock-machine-mapping, intelligent decision-making, multi-constraints, deep learning

## 1. Introduction

As one of the most important underground facilities, tunnel is widely used in railway, highway, water conservancy projects, and urban underground space. Because of its advantages such as high excavation efficiency and strong safety, tunnel boring machine (TBM) has become an important multi-functional equipment in tunnel construction. However, due to its weak adaptability to different geological conditions [1], the selection and adjustment of TBM operating parameters mainly rely on the qualitative understanding of the geological conditions of the rock mass and empirical experience. Much of the selection procedure for the TBM operating parameters is still subject to the experience of the operators and operational manual. Generally, if the operators are experienced and the operational parameters are reasonable, good excavation results can be obtained. However, without a proper understanding and response strategy for the rock formation, it is likely that the selected operating parameters will be unreasonable, making the selected TBM operating parameters hard to adapt to the geological changes, resulting in low excavation speed or utilization rate, abnormal wear of the cutterhead or main bearing, and even serious consequences like TBM blockage [2]. Therefore, a reasonable and effective decision-making method for TBM operation parameters is important to guide the safe and efficient operation of TBMs and is worthy of extensive research.

Studies related to the rational decision making of operating parameters can be divided into two categories, one of which is through the study or analysis of historical excavation data. For example, Zhang et al. [3] introduced an intelligent decision control system by analyzing the operating process of the TBM main driver. Li et al. [4] argued that the intelligent control in TBM tunnelling should aim for the comprehensive optimization of multiple objectives such as safety, operating efficiency, and cost efficiency. By analyzing historical excavation data, Xing et al. [5] proposed an intelligent expert control system, which can adaptively change the driving power of the cutterhead. Duan [6] used the XGboost algorithm to extract changes in operating parameters from historical excavation data, and then predicted operating parameters to assist operators to improve driving efficiency. This kind of method trains the model by studying a large amount of historical data, and then predicts the operating parameters based on the similarity of the rock mass geological conditions. Importantly, this method depends on the quality of historical data and the proportion of outstanding solutions [7]. In other words, the effectiveness of this method depends on high-quality driving data controlled by experienced operators [4].

The other category is based on objective optimization. For example, Sun et al. [8] proposed a TBM operating parameter selection method that considers the tunneling cost or penetration rate and that is constrained by a series of machine data such as cutter life, cutter shape, and torque. Liu et al. [9] established a numerical simulation model of TBM excavation and analyzed the influence of thrust on penetration rate, which can be used as a reference for adjusting TBM thrust. Xue et al. [10] analyzed linear cutting test data and found specific penetration and cutter spacing values, which can obtain the minimum rock-breaking specific energy. Usually, TBM operating parameter optimization is conducted by constructing a correlation between rock parameters and TBM mechanical parameters, based on which the best operating parameters matching the current stratum are obtained to guide efficient TBM tunnelling. The objective optimization method has the advantages of low dependence on historical data and strong geological adaptability.

In the decision-making process of operating parameters, one of the most important challenges is to improve the applicability and rationality of objective optimization methods, which relies on an accurate rock-machine mapping, i.e., the mapping between the rock mass parameters and the TBM tunneling parameters. Many researchers have tried to establish accurate rock-machine mapping through theoretical or empirical methods. Theoretical methods are mainly developed via laboratory tests (e.g., linear cutting tests) and numerical simulations of the rock-breaking process, which empowers the method with a higher adaptability to different rock formation. Empirical models are directly obtained through statistical analysis of field data, and the expression of complex rock-machine relationship has been developed from single factor to multiple factors. For instance, Barton [11] established a QTBM model based on the rock mass classification Q system, which considers the rock mass strength, integrity, density, quartz content, and rock hardness and can accurately predict the penetration rate, utilization rate, and other indexes. In addition, many researchers have proposed a variety of rock-machine mapping models using different methods [12-17]. With the continuous development of computer technology, learning-based techniques like data mining and machine learning have been introduced into the rock-machine mapping with powerful capability in dealing with regression problems with large amounts of data and strong nonlinearity. Different machine learning algorithms has been proposed to simulate the rock-machine mapping and have achieved good results, such as Armaghani et al. [18], Zare et al. [19,20], Mahdeveri et al. [21], Liu et al. [22,23]. These methods have shown a strong ability to solve complex nonlinear relationship between parameters, and as a result the calculating accuracy is relatively higher. Clearly, combining the advantages of the theoretical and empirical methods is an effective way to improve the accuracy of rock-machine mapping, which is beneficial to reduce the negative impact of unstable data quality on data mining, and to modify the learned rules from empirical experience and overcome the inherent shortcoming of data mining [24].

In our previous study, a thrust and torque decision-making method based on multi-objective optimization was proposed, which can theoretically be used to obtain optimized thrust and torque to match the current rock strata, and it has been verified by field tests [7]. To further improve its applicability and rationality in practical data, we tend to introduce the theoretical rule of rock-breaking into the empirical relationship and incorporate constraints to the multi-objective optimization process as an additional accuracy insurance. Thus, in this work, we present a decision-making method for selecting TBM operating parameters based on Constrained multi-objective optimization in two steps. Firstly, we develop a rock-machine mapping of rock parameters and TBM operating parameters dual-driven by data mining and rock-breaking physics. Then, the rock-machine mapping parameters are further used to build multi-objective and constraints for the optimization of TBM operating parameters. The rest of paper is structured as follows. In the next section, we theoretically introduce the proposed method, including the construction process of the dual-driven rock-machine mapping, and the multi-constrained decision-making method. In the third section, we verify the proposed method on the field data in the Second Water Source Channel of Hangzhou, China, and the influence of different rock mass condition on optimal TBM operating parameters is discussed in section four.

## 2. Intelligent decision-making method for TBM operating parameters

This paper puts forward a novel method for selecting optimal TBM operating parameter to achieve safe and economic TBM tunneling. In the following subsections, we firstly introduce a deep neural network (DNN) way to simulate the rock-machine mapping dual-driven by the physical rule and data mining, which can effectively improve the accuracy of rock-machine mappings. Then, we further propose a decision-making method for TBM operating parameters, where the optimization constraints and objectives are designed on the basis of the constructed rock-machine mapping.

### 2.1 Dual-driven rock-machine mapping

In this section, we adopt DNN to simulate the rock-machine mapping model based on physical rule and data mining. It takes TBM operating parameters, rock mass parameters, TBM muck data and cutterhead vibration data as input, and takes cutter life, thrust, torque and the belt conveying volume as output. Here, operating parameters refer to penetration ($P$) and revolution per minute ($RPM$), and rock mass parameters refer to uniaxial compressive strength (*UCS*), rock quality designation (*RQD*) and Cerchar abrasivity index (*CAI*). TBM muck data refer to average particle size, muck geometry, and coarseness index. To improve the mapping accuracy and applicability in different geological conditions, we introduce the known physical rules of rock breaking obtained by numerical simulation and linear cutting test as constraints and priors. How to obtain the physical rule of rock breaking and how to incorporate it into the DNN are two major steps to be concerned with.

2.1.1 Physical rule of rock-breaking by TBM cutter

Linear cutting tests and numerical simulations are two common methods to study the rock-breaking mechanism and physical laws [25-32]. The numerical simulation method is easier to construct a variety of rock mass conditions, and the linear cutting test is closer to the actual rock breaking process. Both have advantages and can be well complementary. In this paper, the numerical simulation method and the linear cutting test are used to simulate and test the rock breaking under different cutting parameters (penetration, cutter spacing, etc.) and rock compressive strength. Specifically, to explore the fragmentation process and the evolution of the TBM cutter load under different working conditions, we use a variety of rock masses with different strengths and set a variety of different penetration and cutter spacing combinations in the linear cutting tests and numerical simulations. In this way, we can obtain the formation of rock breaking slices and the cutter load (normal force and rolling force) under various working conditions. The calculated normal force and rolling force are recorded as $F_N$ and $F_R$, respectively. The relationship between penetration, UCS and the cutter force can be obtained by a polynomial fitting method, as expressed in Equations (1) and (2). Combining the linear analytical relationship between the cutter load and thrust and torque of the cutterhead, we obtain the physical rules between thrust/torque and compressive strength/penetration, as written in Equations (3) and (4) [33], which are denoted as *Th* and *Tor* correlations, respectively.

$$F_N = f_1(UCS, p) \tag{1}$$

$$F_R = f_2(UCS, p) \tag{2}$$

$$Th = N \cdot F_N \tag{3}$$

$$Tor = F_R \cdot \sum_{i=1}^{N} r_i \tag{4}$$

Here *n* is the total number of cutters, and $r_i$ is the distance from the $i^{th}$ cutter to the center cutterhead.

With a cutter spacing of s, as the penetration p increases, the cracks continue to propagate downward and to both sides. When the penetration p increases to a certain value, the cracks generated by adjacent cutters are connected to each other to form rock fragments. In addition, with the same cutter spacing s, a critical penetration exists to create rock fragments. When the penetration p is larger than this value, rocks between cutters can be fragmented successfully. Cho et al. [34] and Moon et al. [32] proved that the critical s/p value is almost a constant under the same rock mass conditions, usually in the range of 10 ~ 20. It can also be established by the regression method as expressed in Equation (5).

$$p_{min} = f(USC, s) \tag{5}$$

This rule is called the CP (Critical Penetration) rule, which will be used to build the constraints of the control parameter optimization model.

2.1.2 Incorporation of physical rule into DNN via constraints

The rock-breaking physics obtained above can be divided into two categories, namely, the equality and the inequality. Specifically, the relationships among the TBM thrust/torque, the rock compressive strength, and penetration are the equality constraints, while the calculated critical penetration, i.e., the minimum penetration required by the formation of rock fragments, is an inequality constraint. The ways to introduce these two types of constraints into the DNN are different. First, the inequality constraints are used to evaluate the rationality of the samples. Samples that meet the inequality constraints are considered normal samples and are trained with larger weights, while samples contrary to the inequality constraints are considered abnormal samples, and their weights are reduced in the DNN training. The weights of abnormal samples can be calculated as

$$Loss_i = \frac{1}{n}(y_i' - y_i)^2 \tag{6}$$

$$Loss_i' = \mu_1 \cdot Loss_i = \frac{1}{n}\mu_1(y_i' - y_i)^2 \tag{7}$$

where $Loss_i$ and $Loss_i'$ represent the contribution of normal or abnormal samples to the $Loss$ function, $n$ is the total number of samples, $y_i'$ and $y_i$ are the network calculated result and the actual result of the $i^{th}$ sample, $\mu_1$ is the weight of the constraint to the sample control, and its value is between 0 and 1. This constraint reduces the weight of the abnormal samples, and increases the influence of normal samples on the physical rule and data mining based rock-machine mapping accordingly.

Unlike inequality constraints, equality constraints represent quantitative relationship between parameters. These quantitative relationships are usually positive references for data mining. The variables calculated by the equality constraints are given weights $\mu_2$, and added into the calculated values to modify it. Eq. 8 is used to introduce the equality constraints into the DNN. Through continuous training, the difference between the physical rule and data mining based mapping and the physics is continuously reduced, which can not only ensure accuracy, but also reduce the risk of overfitting.

$$Loss_i = \frac{1}{n}[(1 - \mu_2)y_i + \mu_2 \cdot y_{ic} - y_i']^2 \tag{8}$$

Here $Loss_i$ is the part of the loss function caused by the $i^{th}$ sample, $n$ is the total number of samples, $y_i$ and $y_{ic}$ represent the values calculated by DNN and the equality constraints, and the $y_i'$ is the actual value.

As mentioned earlier, the rock-breaking physics constraints need to be introduced into the DNN. Three constraints have been built, namely, CP, TH, and TOR models. Among them, the CP model is considered an inequality constraint, while the TH and TOR models are equality constraints. According to the application method, the $Loss$ function under those constraints is calculated by Eq. 9 and Eq. 10 as,

$$\begin{cases} Loss_i = \frac{1}{n}(y_i' - y_i)^2, p_i \geq p_{lim} \\ Loss_i = \frac{1}{n}\mu_1(y_i' - y_i)^2, p_i < p_{lim} \end{cases} \tag{9}$$

$$\begin{cases} E_{Th} = [(1 - \mu_2)Th + \mu_2 \cdot Th_p - Th']^2 \\ E_{Tor} = [(1 - \mu_3)Tor + \mu_3 \cdot Tor_p - Tor']^2 \end{cases} \tag{10}$$

where $p_i$ is the measured penetration of the $i^{th}$ sample, $p_{lim}$ is the critical penetration of rock fragments formation based on the CP rule, $E_{Th}$ and $E_{Tor}$ represent the calculation error of thrust and torque of the cutterhead, respectively, $Th$ and $Tor$ are the measured thrust and torque data of the samples, $Th'$ and $Tor'$ are the calculated results from the physical rule and data mining based mapping, $Thp$ and $Torp$ are the thrust and torque calculated by TH and TOR rules, respectively, and $\mu_1$, $\mu_2$ and $\mu_3$ are the weights of the three constraints.

2.1.3 Establishment of the dual-driven rock-machine mapping

After introducing the rock-breaking physics constraints into the DNN, the network structure and hyper-parameters need to be determined according to the characteristics of the data set. Limited by the total amount of data, the DNN used in this study uses a small number of hidden layers, and the ReLU activation function is used between the layers. In addition, the neuron number of the input layer and output layer are set as the number of input parameters and output targets, respectively. To build a DNN, we need to determine the number of neurons in the hidden layers, the learning rate, the weight of the structure loss, and the weights of physical rule constraints, etc. To this end, the trial-and-error method is used to determine the hyper-parameters of each physical rule. The range and step size of each parameter used in the case study are provided here as an example (see Table 1).

**Table 1. The test ranges of the main hyper-parameters of the DNN**

| Parameters | Testing range | Testing step size | Levels |
| --- | --- | --- | --- |
| Neurons of the 1st hidden layer | $2^8 \sim 2^{10}$ | $\times 2^1$ | 3 |
| Neurons of the 2nd hidden layer | $2^8 \sim 2^{10}$ | $\times 2^1$ | 3 |
| Learning rate $\alpha$ | 0.001~0.005 | 0.001 | 5 |
| Weight of the structure loss $\lambda$ | $10^{-6} \sim 10^{-4}$ | $\times 10^1$ | 3 |
| Constraint weight $\mu_1$ | 0.1~0.9 | 0.1 | 9 |
| Constraint weight $\mu_2$ | 0.1~0.9 | 0.1 | 9 |
| Constraint weight $\mu_3$ | 0.1~0.9 | 0.1 | 9 |

Through trial and error, multiple DNNs with different hyper-parameters and structures are constructed. Then, by comparing the training accuracy of each network model, we can select the network with the highest accuracy to build the physical rule and data mining based rock-machine mapping with physics as constraints.

**2.2 Constrained decision-making method for TBM operating parameters**

2.2.1 Basic process of optimizing TBM operating parameters

In TBM tunneling, how to achieve efficient tunneling and reduce tunneling consumption is one of the most important concerns for the tunnel constructors. TBM excavation efficiency is generally measured by penetration rate, and cutter consumption can be very costly for the tunnel project [35-37]. For example, according to statistics from the Qinling tunnelling project, the cost of cutter consumption accounts for as high as 20% - 30% [38]. In this regard, this paper uses time-related cost affected by penetration rate to characterize tunneling efficiency, and cutter-related cost affected by cutter life to characterize tunneling consumption. A proper setting of TBM penetration and revolution per minute would effectively increase the TBM tunneling speed and cutter life, reducing the TBM operation cost. Thus, we take TBM penetration and revolution per minute as optimization variables and the time-related cost and cutter-related cost are taken as the optimization objective of TBM total excavation cost.

Moreover, keeping the TBM penetrating parameters such as thrust, torque, belt conveying capacity, and critical penetration for the creation of rock fragments within the rated value range is necessary for ensuring TBM tunneling safety, and we take them as constraints in the above optimization. In a word, we develop a decision-making method for optimal TBM operating parameters based on multi-constraints and multi-objective optimization. The basic principles and the process to develop this decision-making method for optimizing TBM operating parameters can be described as follows.

(1) Determine constraints based on the physical rule and data mining based rock-machine mapping and the rated values of TBM operating parameters, which include thrust, torque, belt conveying capacity, and critical penetration. These four constraints ensure that the above TBM indexes are within the rated ranges so the TBM can break rocks efficiently and safely. The detailed procedure of adding multiple constraints into objective optimization can be found in the following part.

(2) Similarly to our previous study [7], the total tunneling cost can be calculated as the sum of time-related costs $C_s$ and cutter-related cost $C_c$, where time-related cost $C_s$ is calculated by multiplying the time by the daily tunneling cost (see equation 12), and the cutter-related cost $C_c$ can be obtained by multiplying the number of cutters by the unit cutter cost (see equation 13).

$$C_t = C_s + C_c \tag{11}$$

$$C_s = \frac{C_1 \cdot L}{\mu \cdot t \cdot PR} \tag{12}$$

$$C_c = \frac{C_2 \cdot L \cdot A}{H_f} \tag{13}$$

Here $C_1$ is the daily cost of TBM construction, primarily including TBM equipment rental cost, staff wages, accommodation, and other expenses; $C_2$ is the single cutter cost; $L$, $A$, $PR$ and $H_f$ are the driving distance, cross-section area, penetration rate, and cutter life, respectively; $\mu$ is the utilization rate of the TBM, i.e., the ratio of daily working time to total time; and $t$ is the total time per day (24h/day). Among these parameters, $L$ and $A$ are constants. Penetration rate and cutter life are functions of operating parameters, where $PR$ is the product of operating parameter $P$ and $RPM$, i.e., $PR = p \cdot rpm$. The relationship 14 represents the dual-driven rock-machine mapping obtained in section 2.1,

$$H_f = f(p, rpm, UCS, RQD, CAI, Mu, Vi) \tag{14}$$

where *Mu* and *Vi* refers to the indicators related to the TBM muck and vibration.

(3) Conduct global optimization search for the objective function established in step 2 using the feasible ranges of the operating parameters obtained in step 1. This method can meet the safety requirement of TBM tunneling while minimizing the excavation cost.

2.2.2 Construction and incorporation of multiple constraints

To optimize the TBM operating parameters, we first need to determine the reasonable ranges of these operating parameters under different rock conditions to ensure normal operation without mechanical damage. If the global optimization of operating parameters (penetration and revolution per minute) is carried out without considering the feasible parameter ranges, two situations may occur. First, the operating parameters may exceed the rated values, which can cause cutter abnormal damage and belt conveyor slipping or blockage. The other situation is that improper operating parameters make it difficult to form uniform rock fragments during rock cutting, which again can cause serious damage to the cutter to shorten its service life. Therefore, considering the constraints and calculating the reasonable ranges of operating parameters are critical to achieve multi-solutions and improve the reliability and stability of the optimization. Furthermore, parameters (e.g., penetration rate) should meet the requirements of TBM rock-breaking to ensure rock fragments are created instead of powder. According to the analysis in section 2.1, the cutter penetration should be larger than the critical penetration corresponding to the current rock mass condition. Otherwise, the rock mass cannot be broken into fragments, which leads to low efficiency.

In addition, we should consider key parameters such as the thrust and torque of the cutterhead that affect the key structures of the TBM (e.g., cutter, cutterhead, and main bearing). When the thrust and torque exceed the rated values, the TBM is overload, which can damage the cutter and the cutterhead. Another issue is that, when the rock mass is fractured or the penetration rate is too large, the amount of muck can significantly increase, leading to belt conveyor slipping and jamming. To summarize, the following are determined as the constraints for selecting TBM operational parameters: rated thrust, rated torque, rated belt conveying capacity, and the critical penetration for the formulation of rock fragments. In general, the critical penetration constraint is used to achieve a good rock-breaking pattern and a high excavation efficiency; and thrust, torque, and belt conveying capacity constraints are used to maintain the integrity of the cutterhead, cutter and belt conveyor. These constraints control the operating parameters such as penetration and cutterhead revolutions per minute.

After the four constraints are determined, the dual-driven rock-machine mapping constructed in section 2.1 is used to give the "contour line" of the penetration and revolution per minute with respect to the extreme value of the constraints. Moreover, considering the rock-machine mapping obtained by learning-based methods is usually implicit expressed, the feasible range of TBM operating parameters under each constraint is calculated in reference to the range and step length of control parameters for actual TBM tunnelling. In this way, assuming that the rock mass parameters, muck geometry, and vibration signal are known, we can assess the combination of operating parameters (i.e., the penetration and cutter revolution per minute), and determine if these parameters meet the constraint requirement against known rated values and the rock-machine mapping. Taking the thrust constraint as an example, we calculate the corresponding thrust value for each operating parameter combination based on the known rock-machine mappings. If the calculated value exceeds the rated thrust, the corresponding operating parameters are not considered to satisfy the thrust constraint. Conversely, when the operating parameter combination yields a thrust value that is lower than the rated value, it is considered to satisfy the constraint. By this procedure, we obtain the combination of the operating parameters satisfying the thrust constraint. The combination of operating parameters satisfying all four constraints simultaneously is a reasonable one, and then these parameters constitute the feasible region of operating parameters, within which the final parameters under the premise of safe tunneling can be obtained by optimization searching.

## 3. Case study

### 3.1 Project overview

In this work, we collected filed data from the Second Water Source Channel of Hangzhou, Zhejiang, China. This project is located in the southwest of Hangzhou, with a total length of 113 km, as shown in Fig.

1. Within the project, the total length of the Jiangnan line is about 26.13km, and the TBM construction section is about 8013m long. According to geological and hydrogeological data along the tunnel, the rock mass has a low strength, poor integrity, a large proportion of class IV rock, abundant groundwater in some areas, and relatively developed karst landforms. The total proportion of class III and class IV rocks along the tunnel is more than 90%, of which class IV rock accounts for more than 52%, and the uniaxial compressive strength of the rock mass is between 30-150 MPa. This project adopted CREC696 double shield TBM produced by China Railway Engineering Equipment Group. According to the TBM operation manual, some basic design parameters of the TBM include: total length (500m), diameter (6m), and the number of total cutters (38).

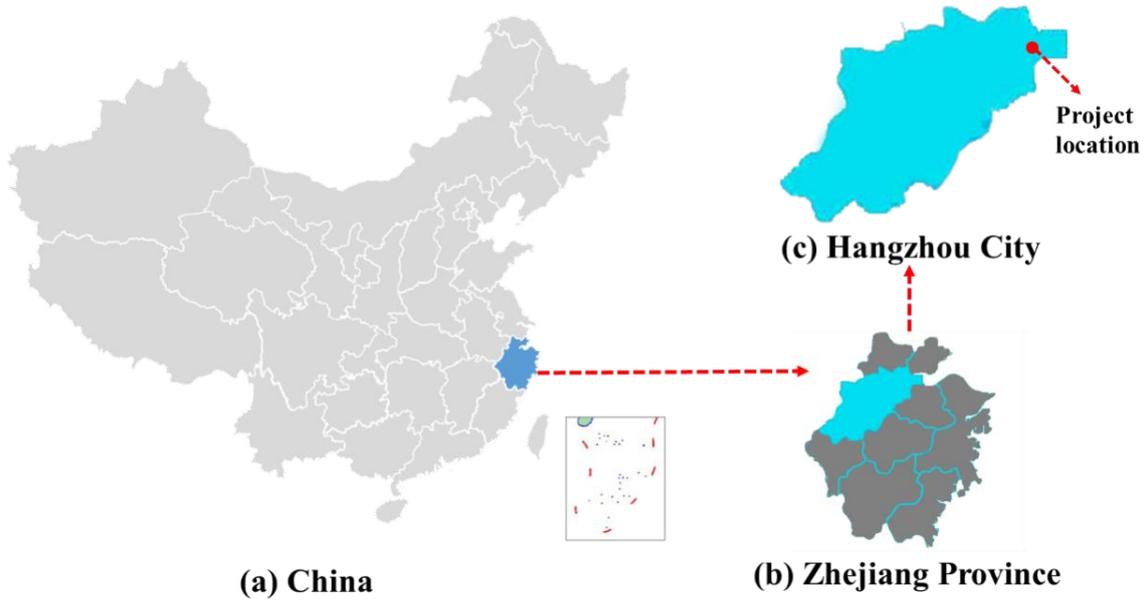

**Fig. 1 The location of the Second Water Source Channel in Hangzhou, Zhejiang, China**

### 3.2 Establishment and verification of dual-driven rock-machine mappings

3.2.1 Field data collection

The main parameters (Key muck index, Key index of the TBM cutterhead vibration) used to establish the rock-machine mapping are collected by the following methods.

Generally, three key indicators are used to characterize muck including average particle size (D), muck geometry, and coarseness index (CI) [39]. The geometry of the muck can reflect the integrity of the rock mass and the wear condition of the TBM cutter. Based on the width-to-length ratio and thickness-to-width ratio of the muck, the geometry can be divided into four types: flat, flat and elongated, elongated, and cubic [39]. Fig. 2 shows the specific classification method of muck geometry. The average particle size and the coarseness index can be obtained by sieving the muck samples. In this work, six sieves with different hole sizes (2.36, 4.75, 9.5, 19, 37.5, and 63mm) were selected for screening, and the two indexes can be obtained by Eq.15 and 16,

$$D = \frac{D_{16}+D_{50}+D_{84}}{3} \tag{15}$$

where $D$ is the average particle size of the muck, $D_{16}$, $D_{50}$ and $D_{84}$ represent the particle size of the muck with a cumulative mass ratio of 16%, 50%, and 84% respectively.

$$CI = \frac{1}{W}\sum_{i=1}^{n}\sum_{j=1}^{i} W_j \tag{16}$$

Here $W$ is the total weight of the muck, $n$ is the total number of sieves used, and $W_j$ is the weight of the residue of j$^{th}$ sieve [39].

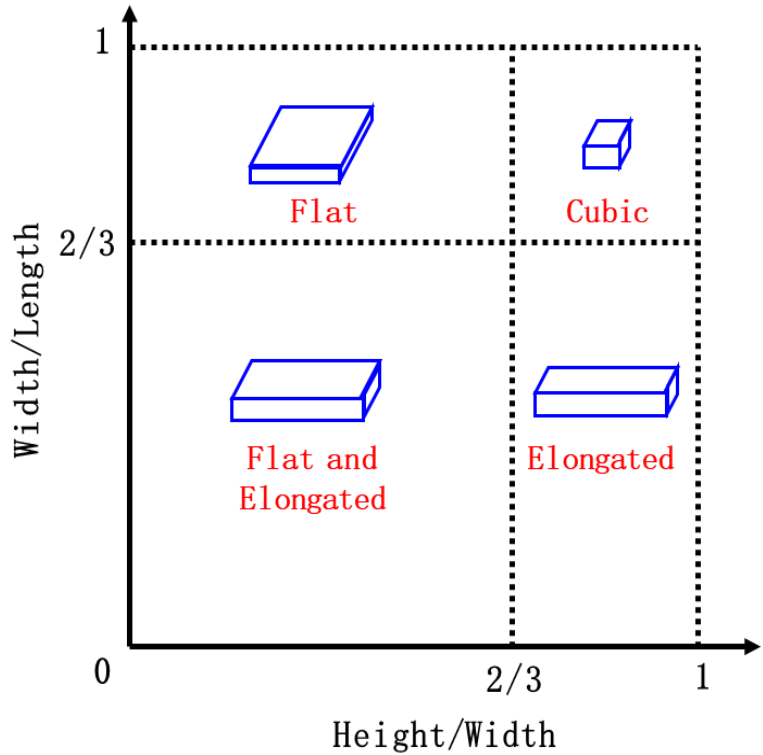

**Fig. 2 Classification method of TBM muck geometry [39,40]**

To collect cutterhead vibration data, a set of cutterhead vibration real-time monitoring system is installed on the TBM [41]. The monitoring system consists of sensors, wireless gateway, remote data transmission subsystem, and data analysis subsystem. A wireless sensor with a protection box is fixed on the back of the cutterhead. The vibration signal of the cutterhead collected by the sensor is transmitted to the wireless gateway through wireless transmission, and then transmitted to the computer for collection and analysis. Using the processing software DataExpir, the key information of the vibration signals can be extracted such as the peak acceleration and the main frequency can be extracted.

Based on the Second Water Source Channel of Hangzhou, we collected rocks and TBM machine parameters through field tests. The collected parameters and their basic statistics information are summarized in Table 2.

| Parameters | Maximum | Minimum | Average | Standard deviation |
|---|---|---|---|---|
| Average size (mm) | 27.69 | 8.86 | 14.87 | 5.72 |
| Coarseness index | 478 | 306 | 383 | 49.3 |
| Peak acceleration (g) | 2.85 | 1.60 | 2.44 | 0.35 |
| Main frequency (HZ) | 116.2 | 110.8 | 113.2 | 1.6 |
| *UCS* (MPa) | 129.3 | 30.3 | 73.0 | 33.1 |
| *RQD* (%) | 79.4 | 11.0 | 39.3 | 19.8 |
| *CAI* (dm) | 4.5 | 1.9 | 3.1 | 0.62 |
| *Th* (kN) | 9127.0 | 1703.3 | 5217.6 | 1821.6 |
| *Tor* (kN·m) | 1824.0 | 286.9 | 791.7 | 332.4 |
| *p* (mm/r) | 15 | 7 | 10.4 | 1.7 |
| *RPM* (rpm) | 6.7 | 5.2 | 4.5 | 0.5 |
| $H_f$ (m³/cutter) | 1116.3 | 176.6 | 538.7 | 263.9 |
| $P_b$ (m³/h) | 583.3 | 486.6 | 536.9 | 24.4 |

Table 2 Data collected from the Second Water Source Channel in Hangzhou

3.2.2 Establishment of dual-driven rock-machine mappings

In order to establish the dual-driven rock-machine mappings, a total of 40 sets of linear cutting test data and 216 sets of numerically simulated data are adopted to simulate the physical rule of rock-breaking. In the linear cutting test, as shown in Fig. 3 and 4, granite and sandstone with compressive strength in the range of 60-180 MPa are tested under different penetrations [26,27]. In addition to laboratory tests, Particle Flow Code (PFC) is used to conduct the rock-breaking numerical simulations and the cutter is simulated to be the 17-inch constant cross-section cutter commonly used in TBM engineering [42]. According to the different moving directions of the cutter relative to the rock surface, two types of rock samples, namely indentation model and the cutting model (see Fig. 5), are used for the calculation of cutter normal force and rolling force, respectively. Moreover, to further explore the fragmentation process and the evolution of the cutter load, we use a variety of rock masses with different strengths and set a variety of different penetration and cutter spacing combinations. As shown in Table 3, a total of 6 rock mass conditions with a compressive strength of 50 - 300 MPa (in 50 MPa steps) are established. Penetration and cutting are simulated by 1-9 mm penetration (in 1 mm steps) and 60-90 mm cutter spacing (in 10 mm steps), respectively.

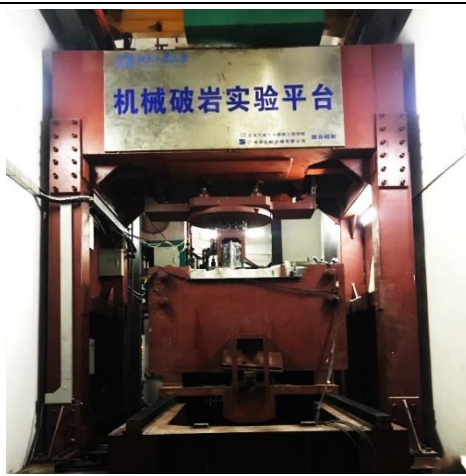 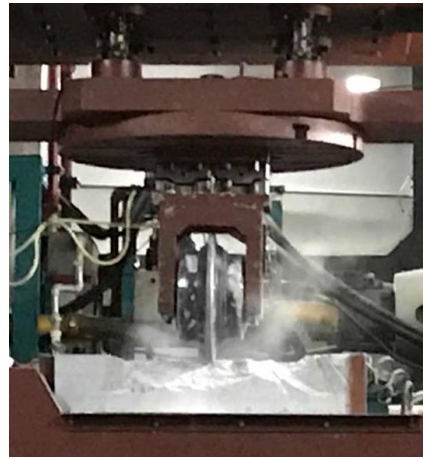

(a)            (b)

**Fig.3 The linear cutting test experimental equipment from Beijing University of Technology, China. (a) Picture of the total equipment (b) Picture of the testing procedure [43]**

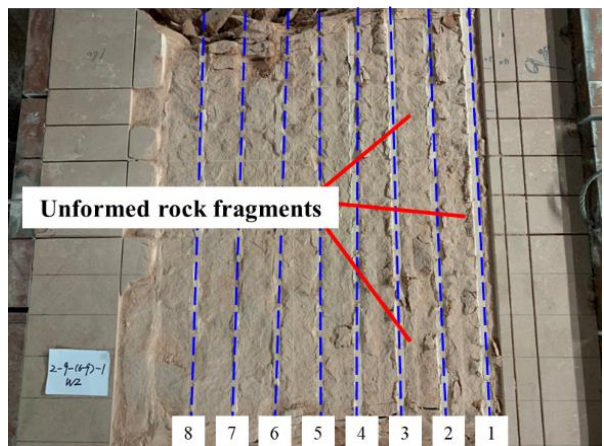 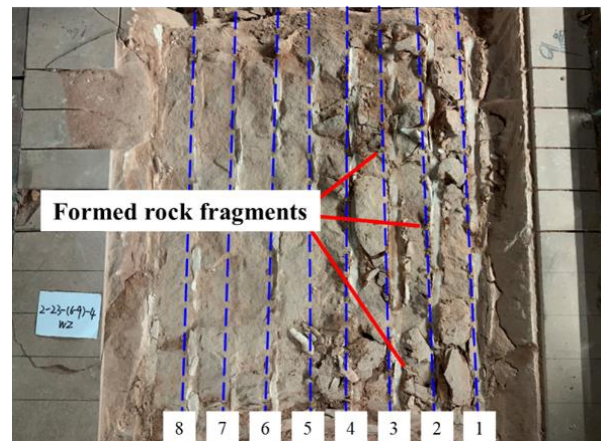

(a)            (b)

**Fig. 4 The formation condition of the red sandstone muck under different penetration (a) penetration =1mm, cutter spacing =70mm (b) penetration =4mm, cutter spacing =70mm**

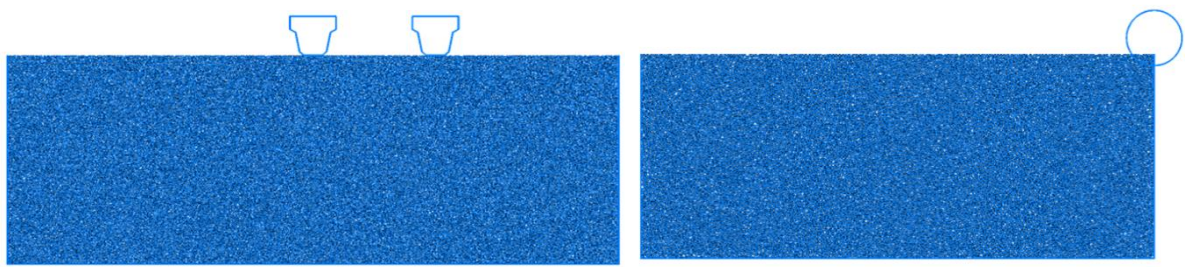

(a)            (b)

**Fig.5 Numerical simulation models of rock-breaking by TBM cutters. (a) Indention model of size 40 cm×16 cm, (b) cutting model of size 140 cm×56 cm [43].**

Table 3 The main parameters used in the numerical simulation

| Parameter | Minimum | Maximum | Step size |
|---|---|---|---|
| Penetration (mm/r) | 1 | 9 | 1 |
| Cutter spacing (mm) | 60 | 90 | 10 |
| UCS (MPa) | 50 | 300 | 50 |

On the basis of abovementioned data, the physical rule of TBM cutter rock-breaking can be obtained following the instructions in section 2.1. In terms of the cutter load law, the 40 sets of linear cutting test data and 54 sets of numerically simulated data with the same cutter spacing are used to fit a polynomial equation (see Fig. 6). In this way, the equation 1 and 2 can turn to equation 17 and 18 and the mapping relationship between cutter thrust and torque, and compressive strength and penetration can be subsequently obtained via equation 3 and 4. As for the formation rules of rock fragments, all of the linear cutting test data and numerically simulated data are used to fit the critical s/p value and the compressive strength, which results in the CP rule shown in equation 19.

$$F_N = f_1(UCS, p) = -1.5 \times 10^{-3} \cdot UCS^2 + 0.26 \cdot UCS \cdot p - 0.74 \cdot p^2 + 0.79 \cdot UCS + 0.6 \cdot p + 2.72 \quad (17)$$

$$F_R = f_2(UCS, p) = -1.44 \times 10^{-4} \cdot UCS^2 + 0.05 \cdot UCS \cdot p - 0.12 \cdot p^2 + 0.01 \cdot UCS + 0.13 \cdot p - 1.8 \quad (18)$$

$$p_{min} = f(USC, s) = \frac{s}{-0.0359 \cdot UCS + 21.1} \quad (19)$$

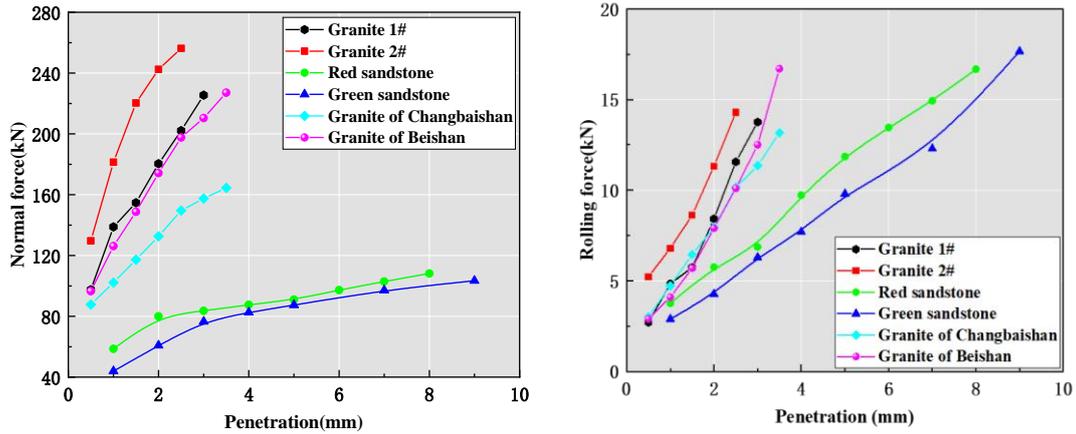

Fig. 6 The cutter data as a function of penetration in the linear cutting tests. (a) normal force, (b) rolling force.

Table 4. Main hyper-parameters used in the DNN

| Parameter | Value |
|---|---|
| Neurons of input layer | 9 |
| Neurons of the 1st hidden layer | $2^{10}$ |
| Neurons of the 2nd hidden layer | $2^{10}$ |
| Neurons of input layer | 4 |
| Learning rate α | 0.001 |
| Weight of the structure loss $\lambda$ | $10^{-5}$ |
| Constraint weight $\mu_1$ | 0.2 |
| Constraint weight $\mu_2$ | 0.1 |
| Constraint weight $\mu_3$ | 0.2 |

After constructing the constraints on the basis of these physical rule, 306 field samples collected from the Second Water Source Channel of Hangzhou, China are used for DNN training and verification. Specifically, 256 samples are randomly divided into the training data set and the other 50 samples are used for testing. By following the deep learning algorithm described in section 2.1, a physical rule and data mining dual-driven rock-machine mapping is developed with the objectives of cutter life, thrust, torque and belt conveying volume. Table 4 lists the hyper-parameters used in the algorithm. With the trained DNN, we can input the TBM operation parameter, muck information and vibration signal, and obtain the calculated cutter life, thrust, torque, and belt conveying capacity of the TBM.

3.2.3 Verification of dual-driven rock-machine mappings

To verify the positive effect of the physics constraints on the dual-driven rock-machine mapping, DNNs with the same structure and hyper-parameters are trained to build pure empirical mapping without physics constraints for the same target. The empirical mapping is then applied to the same 50 sets of field samples, and the improvement effect of the physics constraints on the mapping is found by comparing the error between the calculated results and the measured results, as shown in Figs. 7-10.

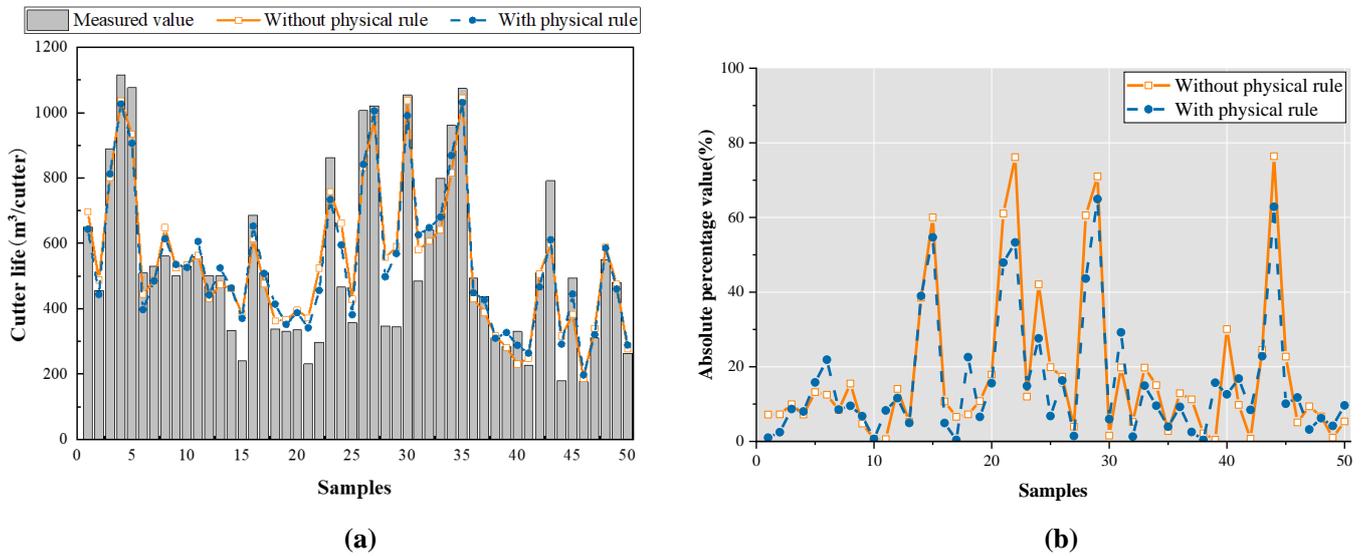

**Fig. 7 The comparison of the cutter life calculated by the physical rule and data mining based rock-mapping and the DNN without physics constraints. (a) the calculation comparison of the cutter life by different methods (b) the comparison of the calculated percentage error of the cutter life**

As shown in Fig.7, the calculated results of the cutter life mapping are in good agreement with the measured results with similar trend. The mean absolute percentage error (MAPE) is only 15.8%, and the coefficient of determination $R^2$ is 0.72, indicating that the calculated results are close to the measured results. Compared with the empirical mapping (MAPE of 18.1%), the accuracy of the physical rule and data mining based mapping is improved, and the MAPE is reduced by 2.3%. Fig.7 (b) illustrates that the percentage error of the two mapping results has several peaks, corresponding to the samples with larger calculation error, such as #15, #21, #22, #28, #29, and #44. The calculation accuracy of the large error samples has been significantly improved, the error reduction is more than 3%, and the reduction in the sample #22 is as high as 20%.

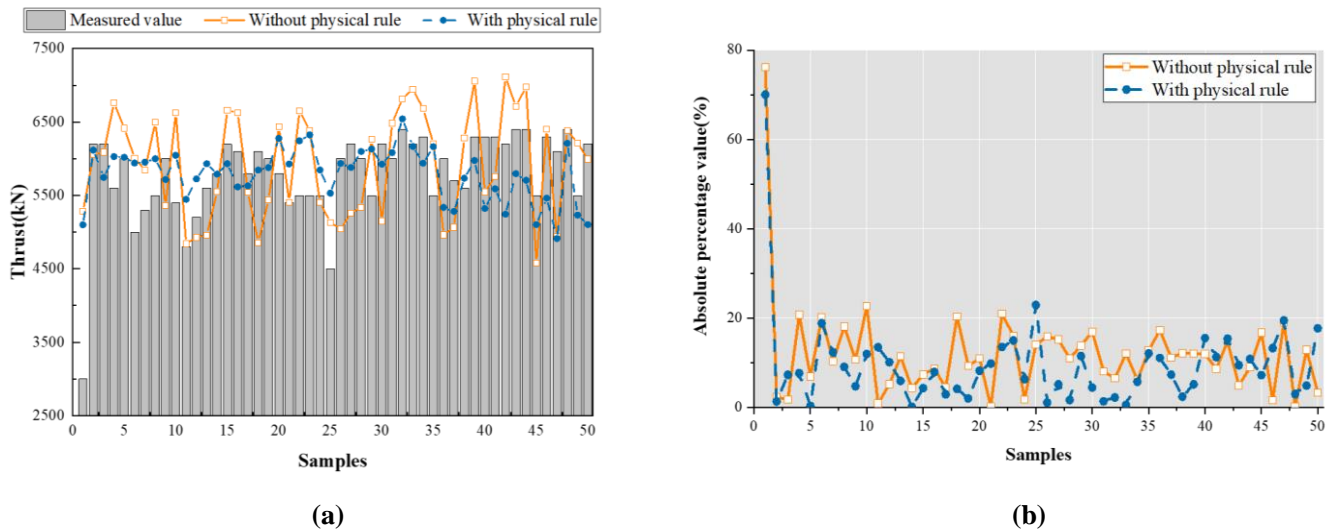

(a)          (b)

**Fig. 8 The comparison of the TBM thrust calculated by the physical rule and data mining based mapping and the DNN without physics constraints. (b) the calculation of the TBM thrust by different methods (b) the comparison of the calculated percentage error of the TBM thrust**

As shown in Fig.8, the MAPE of the thrust calculation error is 9.4%, and the relative percentage error of all samples is within 20% except for the #1 sample. The average error of the thrust calculation based on empirical mapping is close to that of the physical rule and data mining based mapping. Specifically, the MAPE of the empirical mapping is 12.0%, while the value of the physical rule and data mining based mapping is 9.4%. Especially for the #1 sample, the measured thrust value is only 3000kN, which is significantly different from other samples. Therefore, the calculated results of empirical DNNs and physical rule and data mining based mapping have large deviations compared with other samples, which are 76.1% and 69.0% respectively. It proves that for special outliers, the calculated result of the physical rule and data mining based mapping is more accurate. Except for sample #1, the MAPEs of empirical DNNs and the physical rule and data mining based mapping are within 21.4% and 20.1%, respectively, indicating that both models yield accurate thrust results for most samples.

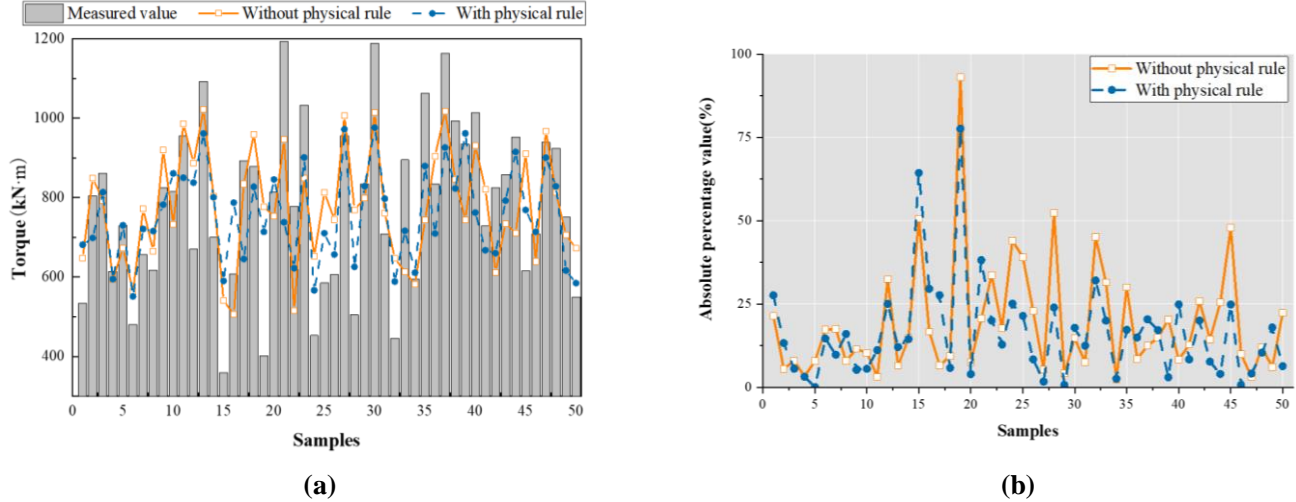

(a)          (b)

**Fig. 9** The comparison of the torque calculated by the physical rule and data mining based mapping and the DNN without physics constraints. (a) the calculation of the torque by different methods (b) the comparison of the calculated percentage error of the torque

Fig.9 shows that the MAPE of the torque calculation is 16.2%, and the relative percentage error of the torque calculation results is within 40%, except for samples #15 and #19. The calculated MAPEs of the empirical mapping and the physical rule and data mining based mapping for torque are 19.3% and 16.2%, respectively, and their floating ranges are 402.2 ~ 1038.8 kN·m and 569.0 ~ 976.9 kN·m, respectively. Compared with the empirical mapping, the calculation accuracy of the physical rule and data mining based mapping is slightly improved. For the samples with large errors, such as samples #19, #28, #32, and #45, the physical rule and data mining based mapping can significantly improve the results. For example, the calculation error of sample #19 is reduced from 92.4% to 76.4%.

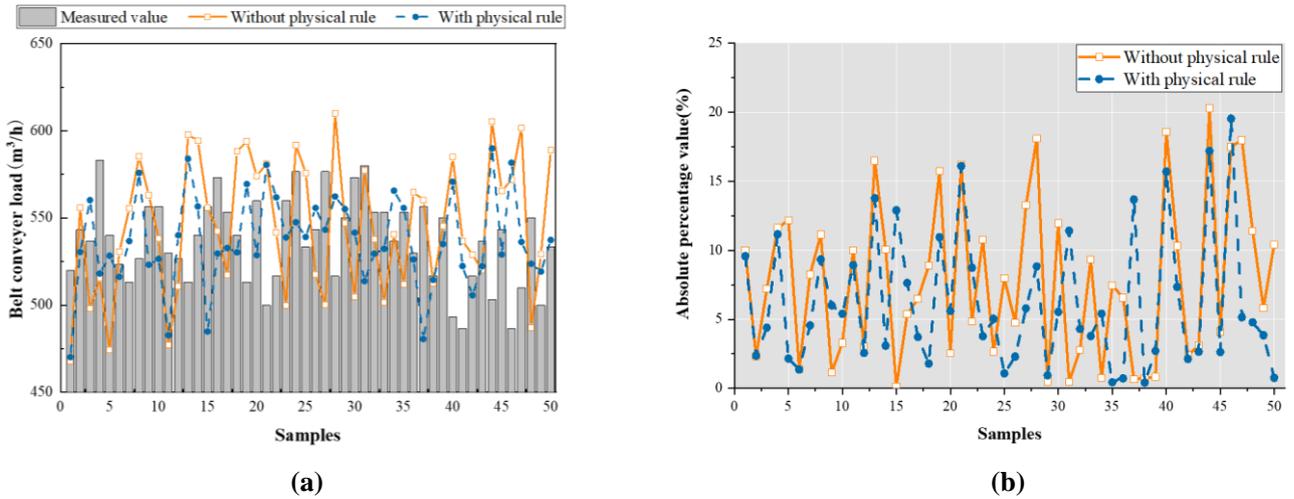

(a)　　　　　　　　　　　　　　　(b)

**Fig. 10** The comparison of the belt conveying volume calculated by the physical rule and data mining based mapping and the DNN without physics constraints. (a) the calculation of the belt conveying volume by different methods (b) the comparison of the calculated percentage error of the belt conveying volume

Compared with cutter life, thrust and torque, the data of belt conveying volume is relatively stable, and the calculated results are better (Fig.10). The MAPEs from the empirical DNN and the physical rule and data mining based mapping are 7.8% and 6.2%, respectively. The sample error distribution is relatively uniform, and the maximum error does not exceed 18.4% and 20.2%, respectively. Unlike the physical rule and data mining based mapping, the maximum error of the empirical DNN is 66 $m^3$/h (sample #45).

By comparing the calculated results of 50 sets of random samples, we find that the calculation error of the physical rule and data mining based rock-machine mapping is slightly lower than that of the empirical DNN, especially for the abnormal samples such as the thrust value of sample #1 and torque value of samples #19 and #28. Our results show that the calculation accuracy of the physical rule and data mining based mapping is higher than that of the conventional rock-machine mapping. Therefore, these mapping methods are used to build the objective function and constraints for the TBM operating parameter optimization.

### 3.3 Development and verification of the decision-making model based on filed data

In this section, we will introduce the development process of the decision-making model for TBM operating parameters based on field data. Because decision-making models are based on field rock conditions and TBM parameters, each model is only applicable to a specific field project. However, these different models can be developed in the same way, as described in section 2. In the following, we show the development process of the decision-making model of TBM operating parameters for the Second Water Source Channel of Hangzhou.

3.3.1 Establishment of the decision-making model based on filed data

With the developed rock-machine mapping in the above section, the rated values of the key machine parameters (such as thrust, torque, and belt conveying capacity) are needed to define the constraints for the decision-making model, which are listed in Table 5. As mentioned in Section 2.2, the feasible range of parameter can be obtained in reference to the range of the control parameters for actual TBM boring (e.g., the speed range is 0-10 rpm with a step size of 0.1 rpm, and the penetration range is 0-16 mm/r with a step size of 1 mm/r). Taking UCS value of 100 MPa, RQD value of 80% and CAI value of 3.5 dm as an example, through the calculation of rock-machine mapping model, the range of control parameters satisfying each constraint is shown and the final feasible range can be obtained via intersection in Fig. 11.

Table 5 The rated values of the TBM in the Second Water Source Channel of Hangzhou

| Key TBM parameters | Thrust (kN) | Torque (kN·m) | Belt conveying capacity (m$^3$/h) |
| --- | --- | --- | --- |
| Rated value | 30000 | 3780 | 600 |

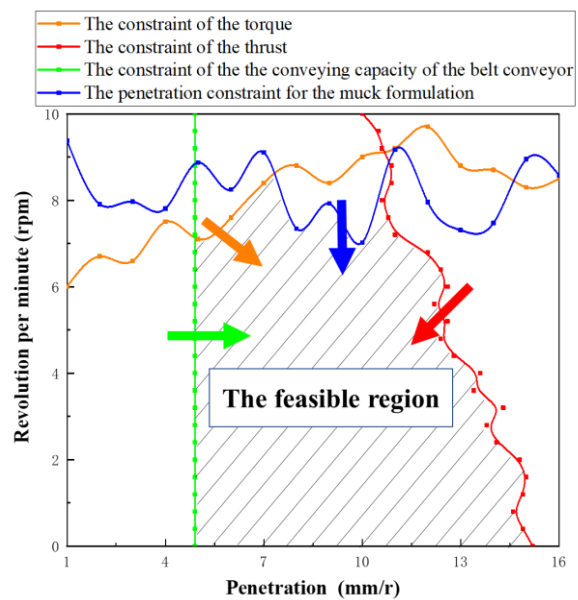

Fig. 11 The constraints and feasible region of the optimized operating parameters for the TBM under given rock mass parameters (The arrows represent the direction of the constraints)

Thrust and torque are usually lower than 50% of the rated values under normal excavation conditions. Therefore, to ensure excavation safety, we use 0.4 as the safety factor for thrust and torque during the development of the decision model. i.e., the thrust and torque are 12000kN and 1512kN·m, respectively. Next, we determine the ranges of the operating parameters and the search steps based on the field historic data, as provided in Table 6. Finally, we construct the decision-making model based on the rated values of the machine parameters, the developed rock-machine mapping, and the ranges and search steps of the operating parameters. Using this model, the optimal operating parameters can be calculated from the series of inputs.

**Table 6 The ranges and search steps of the operating parameters**

|  | Revolution per minute (rpm) | Penetration (mm/r) |
|---|---|---|
| Range | [0, 10] | [0, 16] |
| Search step | 0.1 | 1 |

3.3.2 Verification of the decision-making model by excavation tests

To verify the decision-making method and model of TBM operating parameters, field excavation tests with a length of about 400 m were carried out in the Second Water Source Channel of Hangzhou. In our previous study, field excavation tests were also conducted in the same tunnel, but in a different area [7]. Two major differences exist between the previous and the current decision-making methods and models. First, the optimization objectives are different. The objective in the previous study was to optimize the thrust and torque, but in this work, revolution per minute and penetration are considered the decision-making objectives. Second, we determine several constraints on the TBM operating parameters to control the optimization and ensure safe excavation in this work, while the previous study did not incorporate constraints on the key machine parameter during optimization.

The rock mass in the excavating section is mainly composed of class III silty mudstone, argillaceous siltstone, and fine siltstone, without large-scale faults, fracture zones and other geological structures. The geological conditions of all rock strata are similar without significant changes in structures and in properties. Through fieldwork, we obtained rock parameters such as strength, integrity, and abrasiveness in this excavating section. Combining with the real-time muck data and vibration signal during the excavation process, we obtained the optimal penetration and revolution per minute from the established operating parameter optimization model. Subsequently, operating parameters obtained using the new decision-making model were used in the section from 0+288 to 0+218 (70m) and the section from 0+113 to 0+6 (107m) during tunneling. The results were compared with other sections of the tunnel where operating parameters were selected by the TBM operator. We compared the penetration rate and the cutter life of these two different types of sections and then calculated the time-related cost, cutter-related cost, and total excavation cost. Table 7 lists the range of each test section and the corresponding method for selecting TBM operating parameters.

To verify the optimization model for TBM operating parameters proposed in this work, input parameters required to build the model include *UCS*, *RQD*, *CAI*, muck geometry, and vibration signal. Among these parameters, vibration signal and muck geometry are influenced by the selected tunneling parameters, TBM construction status, and other factors, which cannot be obtained in advance and must be collected in real time during the tunneling process. Cutterhead vibration signal can be real-time monitored and recorded, and the characteristics of muck can be quickly recorded after rock muck being collected in the field. Unlike muck geometry and cutterhead vibration signal, *UCS*, *RQD*, and *CAI* take a long time to obtain, so they cannot be used for real-time acquisition and for operating parameter decision-making and must be known in advance. In addition, *UCS*, *RQD*, and *CAI* of this section are taken as constants because the geological conditions of the testing section are relatively homogenous, and the variation of rock properties is small. Six sets of *UCS*, *RQD*, and *CAI* values were collected in the range of 0 + 420 ~ 0 + 360 (60m). The mean value is taken as the rock mass parameters of the section and used as the input parameters to optimize the operating parameters, which are provided in Table 8.

**Table 7 Range of each test section and corresponding method for selecting TBM operating parameters**

| Section number | Range of excavation position(m) | Section length (m) | Decision-making method |
|---|---|---|---|
| I | 0+352 – 0+288 | 64 | Operator control |
| II | 0+288 – 0+218 | 70 | Model decision |
| III | 0+218 – 0+113 | 105 | Operator control |
| IV | 0+113 – 0+6 | 107 | Model decision |
| V | 0+6 – 0-48 | 54 | Operator control |

**Table 8 The parameters of the rock mass in the excavation testing sections**

| Parameter | *UCS* | *RQD* | *CAI* |
|---|---|---|---|
| Testing result | 82MPa | 31.2% | 3.67dm |

After obtaining the data needed for the optimization of TBM operating parameters (Table 8), we collected and measured rock muck during the idle time intervals between adjacent excavation cycles, and used the newly developed decision-making model to optimize operating parameters. With the required rock parameters available and the fast turn-around of the optimization model, the optimal TBM operating parameters for each excavation cycle can be obtained quickly and adjusted accordingly., The operating parameters obtained by the optimization model vary depending on the excavation position along the tunnel, as shown in Fig. 12 and Fig.13.

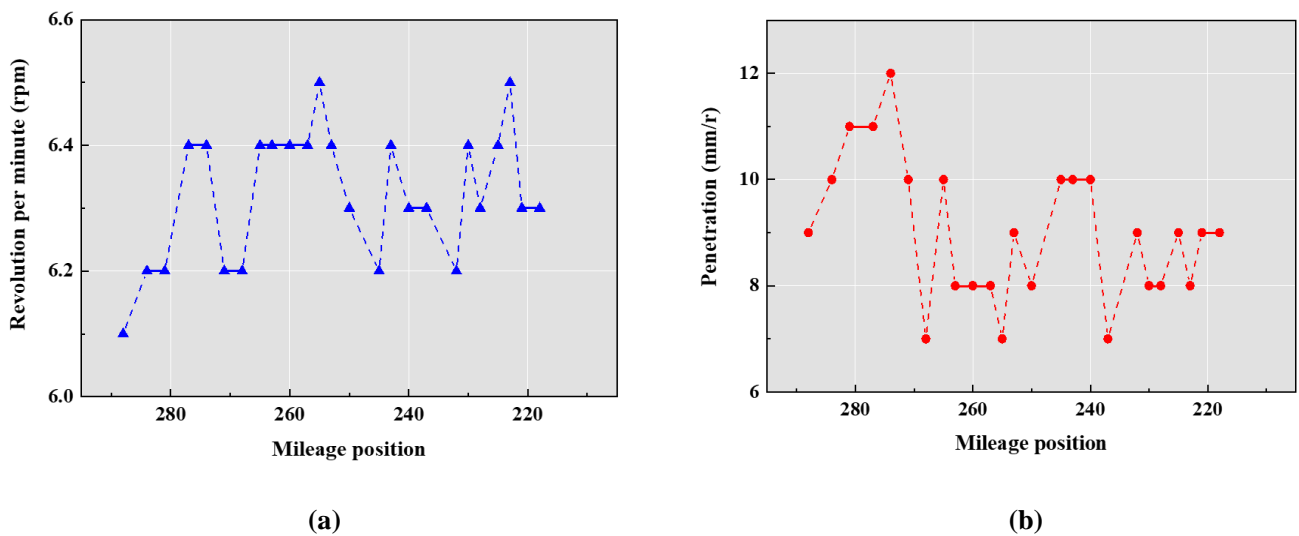

(a)                                     (b)

**Fig. 12 The optimized results of the TBM operating parameters in section II (0+288 – 0+218, 70m). (a) revolution per minute as a function of excavation position from the optimization model (b) penetration as a function of excavation position from the optimization model**

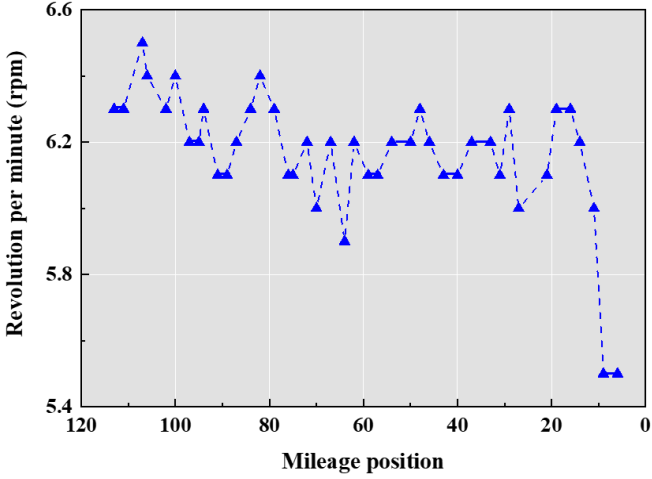
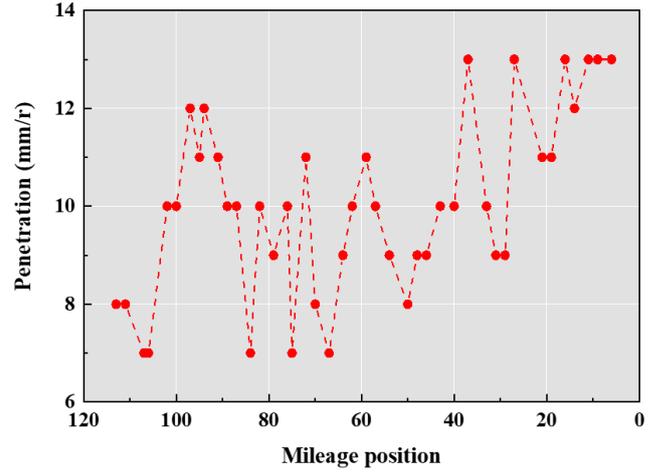

(a)            (b)

Fig. 13 The optimized results of the TBM operating parameters in section IV (0+113 – 0+6, 107m) (a) revolution per minute as a function of excavation position from the optimization model (b) penetration as a function of excavation position from the optimization model

    The optimization model for TBM operating parameters proposed in this paper is based on the total tunneling cost and consists of time-related cost and cutter-related cost. These two parts are directly affected by the penetration rate and cutter life. The higher the penetration rate and the longer the cutter life, the lower the total tunneling cost. To assess the advantages of this optimization model, we compare the penetration rate, cutting life, and weighted averages between the model-optimized sections and the operator-controlled sections. The weights of the penetration rate and the cutter life are the total length of each section. The calculation of the penetration rate and the cutter life can be expressed as

$$\overline{PR} = \frac{1}{L}\sum_{i=1}^{n} L_i \cdot PR_i \tag{20}$$

$$\overline{Hf} = \frac{1}{L}\sum_{i=1}^{n} L_i \cdot Hf_i \tag{21}$$

    As shown in Fig. 14 and Fig. 15, the penetration rates obtained from the optimization model in section II and section IV are 57.7mm/min and 62.0mm/min, respectively. The average penetration rate in these two sections is 60.3mm/min, which is 12% higher than the average penetration rate in the operator-controlled sections, 53.8mm/min (section I: 56.0mm/min; section III: 49.2mm/min; section V: 60.4mm/min).Moreover, the cutter life in the optimized section II reaches the maximum value of 619.4m$^3$/cutter among all sections, and an average cutter life of 556.9m$^3$/cutter in the model-optimized sections is close to the average in the operator-controlled sections of 553.8m$^3$/cutter, with a difference of 0.5%. These results demonstrate that the operating parameters obtained by the optimization model can improve the penetration rate and cutter life, especially the penetration rate.

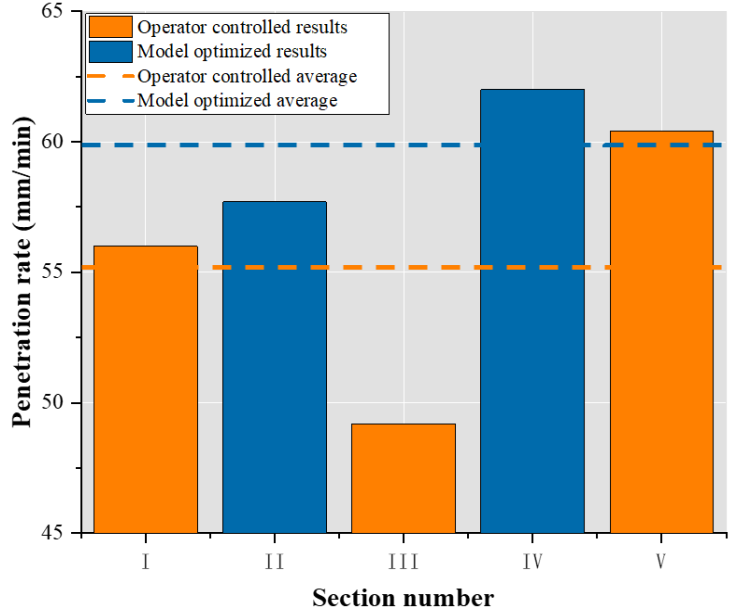

**Fig. 14 Comparison of the penetration rate in model-optimized sections and operator-controlled sections**

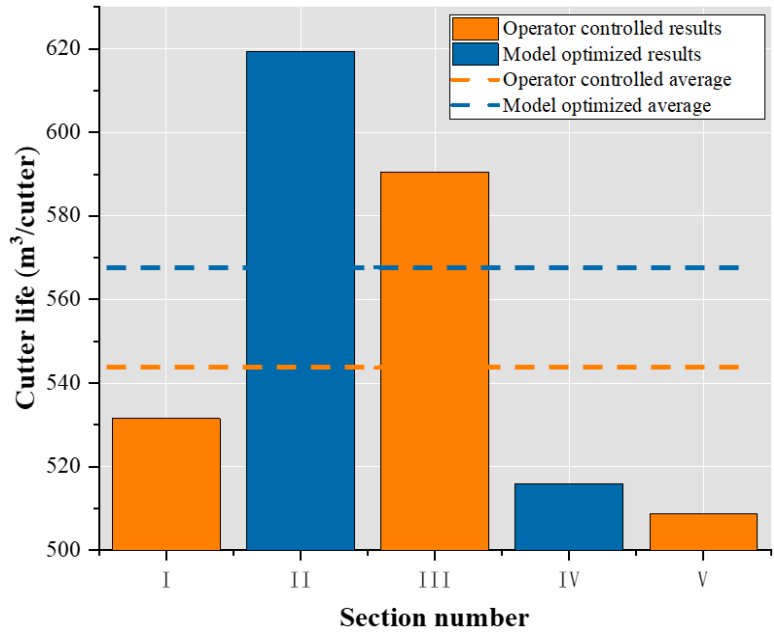

**Fig. 15 Comparison of the cutter life in the model-optimized sections and operator-controlled sections**

As stated in Equation (11), the total tunnel cost is obtained by adding the time-related cost and the cutter-related cost. Consequently, the total tunneling cost is also reduced due to increased penetration rate and cutter life. Fig. 16 shows the total excavation cost and its average in all sections. Similar to the penetration rate and cutter life, the average cost is calculated using a weighted average method and the weighting factor is equal to the total length of the section, expressed as,

$$\overline{C_c} = \frac{1}{L}\sum_{i=1}^{n} L_i \cdot C_{ci} \tag{22}$$

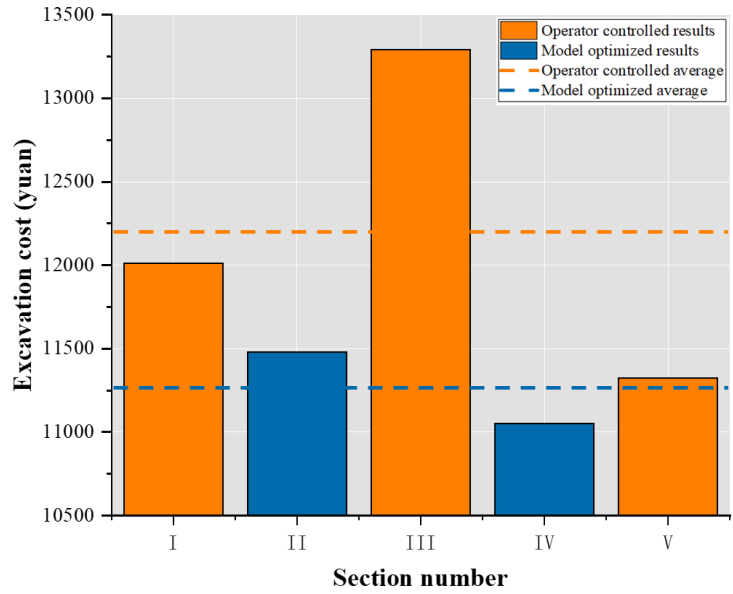

**Fig. 16 Comparison of the total cost in the model-optimized and operator-controlled sections**

Fig.16 shows that, after optimizing the operating parameters by the decision-making model, the average total excavation cost in sections II and IV is 11220.5 yuan, which is 10.0% lower than that in the operator-controlled sections (12448.1 yuan). These results indicate that the operating parameters obtained by the decision-making model are feasible and effective.

## 4. Discussion

In this paper, 50 samples collected from the Second Water Source Channel of Hangzhou are used to verified the proposed decision-making model. However, limited geological conditions are collected in the single projects, and the applicability of the developed model in different rock conditions should be further discussed. For this purpose, taking the rock mass parameters as example, the developed model is applied on multiple areas assumed with different combinations of *USC*, *RQD* and *CAI* value, and the optimal operating parameters are deducted. The ranges and steps of the rock mass parameters are given in Table 9.

**Table 9 The ranges and steps of each rock mass parameter**

| Parameter (Unit) | Minimum | Maximum | Step |
| --- | --- | --- | --- |
| UCS (MPa) | 50 | 300 | 50 |
| RQD (%) | 20 | 100 | 20 |
| CAI (dm) | 2 | 5 | 1 |

As the Table 9 shows, there are 6, 5 and 4 factor levels of *UCS*, *RQD* and *CAI*, and each combination of the rock mass parameters are corresponding to a single decision-making result. Therefore, for a single parameter, there are more than one optimal result. For example, under the condition of the same *UCS*, there are totally 20 optimal results corresponding to the different *RQD* and *CAI*. Similarly, there are 24 results and 30 results under a same *RQD* and *CAI*. Therefore, box diagram instead of scatter or line diagram are used to express the deduction results, which is shown in Fig. 17.

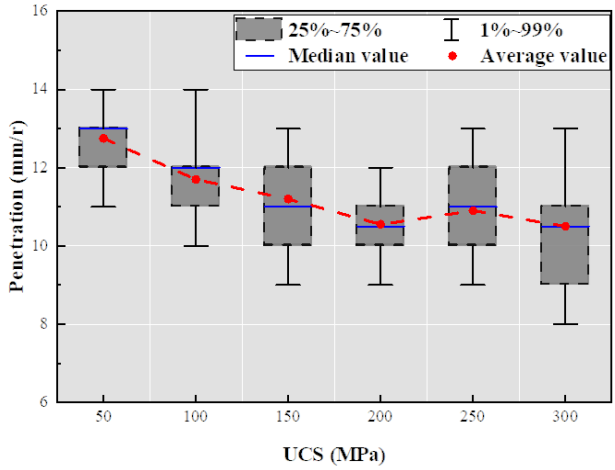
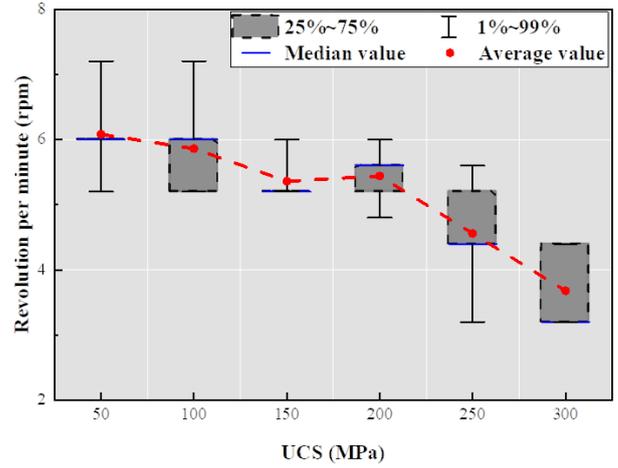

(a)                          (b)

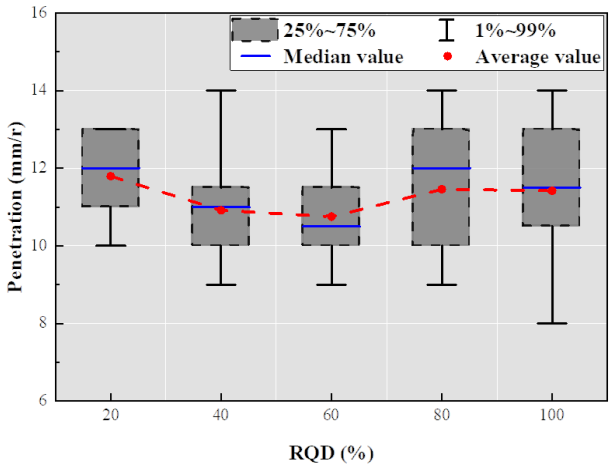
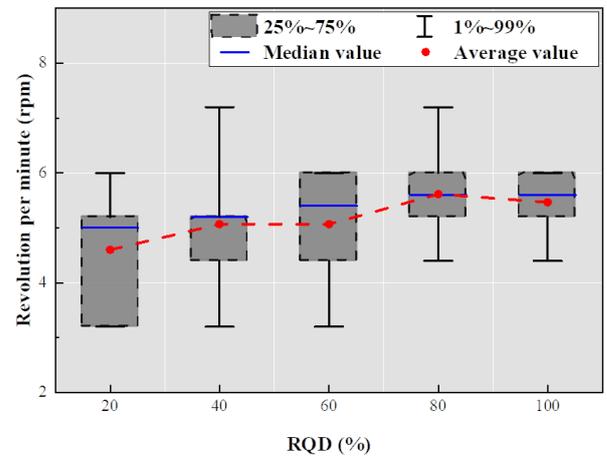

(c)                          (d)

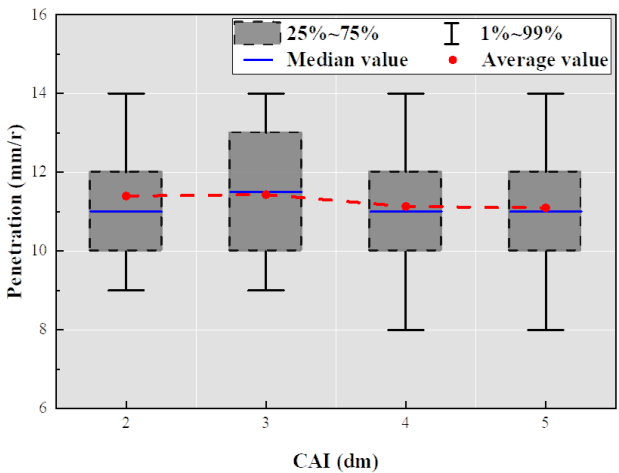
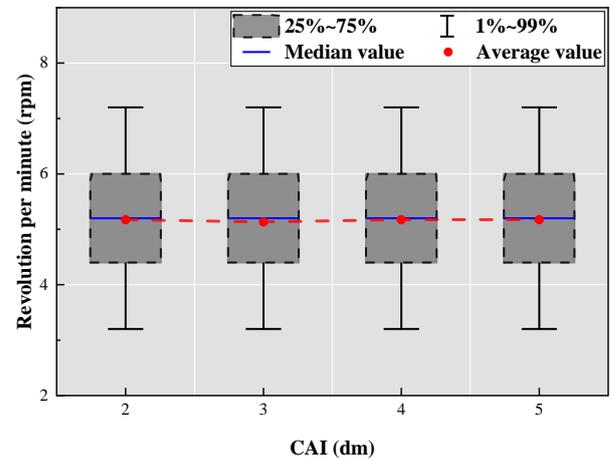

(e)                          (f)

**Fig. 17 Deduction results with multiple combinations of rock mass parameters (The red dotted line represents the average connecting line). (a) statistic law between optimal penetration and UCS (b) statistic law between optimal revolution per minute and UCS (c) statistic law between optimal penetration and RQD (d) statistic law between optimal revolution per minute and RQD (e) statistic law between optimal penetration and CAI (f) statistic law between optimal revolution per minute and CAI**

  The Fig.17-(a) and (b) shows the optimal penetration and revolution per minute changing with different UCS. Generally, the optimal operating parameters decrease with the increasing of UCS. In addition, the range of the optimal operating parameters under more than 200MPa are larger than the one lower than 100MPa, which means that, under high rock strength condition, factors except strength such as RQD and CAI have significant influence on the selection of operating parameters. The Fig.17-(c) and (d) shows the optimal penetration and revolution per minute changing with different RQD. The optimal revolution per minute increases obviously with the increasing of RQD. As for the penetration, there is not a monotonic relationship with RQD, and the average penetration reaches its minimum of 10.6mm/r when the RQD closed to about 60%. The Fig.17-(e) and (f) shows the optimal penetration and revolution per minute changing with different CAI. The average optimal penetrations and revolution per minutes are floating in a small range, which is 11mm/r-11.5mm/r and 5.0-5.2rpm. Compared with UCS and RQD, there is a relatively small influence of CAI on the operating parameters.

The above orthogonal test indicates that the proposed method preserves a certain generalizability. The decision-making model developed in a single project can be applied in different rock condition and present a valuable reference for the changing trend of the TBM operating parameter. How to quantitively determine the optimal TBM operating parameter adjustment with respect to the different rock conditions are an important topic worth further research.

## 5. Conclusion

A decision-making method was proposed to optimize TBM operating parameters with multiple constraints and objectives. Numerical simulation and linear cutting tests were used to explore the formation rules of rock fragments and the evolution of the cutter load, and the corresponding mapping relationship was established. Subsequently, the rock-breaking physics were introduced into the DNN as constraints, and an integrated physical rule and data-mining based rock-machine mapping model was developed. Based on the rock-machine mapping, we proposed a decision-making method and model for optimizing TBM operating parameters, where the total tunneling cost consisting of the penetration rate and cutter life was taken as the objective function, with the thrust, torque, and belt conveying capacity as the constraints. Under these constraints, the searching area of revolution per minute and penetration can be determined and their optimal value can be obtained corresponding to the extremum of the objective. The proposed method is verified based on the field data from the Second Water Source Channel of Hangzhou, China. Compared with the operator-controlled sections, the total cost in the model-optimized sections was decreased by 10.0%, the average penetration rate and cutter life were increased by 12.0% and 0.5%, respectively. These improvements have demonstrated that our developed model is effective in ensuring safe excavation and improving cost efficiency. In the end, the applicability of the developed model in different rock conditions is demonstrated, leading to a potential development direction for the proposed method in the future.

## Acknowledgement

This research was supported by the National Natural Science Fund (No. 52021005, 51922067, 51991391), the Key Research and Development Plan of Shandong Province (No. 2020ZLYS01), and Taishan Scholars Program of Shandong Province of China (tsqn201909003). This support is gratefully acknowledged. The corresponding author thanks Zhejiang Tunnel Engineering Group Co., Ltd, and China Railway Engineering Group Co., Ltd for their support.